\documentclass{article}

\usepackage{microtype}
\usepackage{graphicx}
\usepackage{booktabs} 
\usepackage{amssymb}
\usepackage{amsmath}
\usepackage{subcaption}
\usepackage{bbm}
\usepackage{enumitem}
	
\usepackage{hyperref}

\usepackage[accepted]{icml2018}
\icmltitlerunning{Reinforcement Learning from Imperfect Demonstrations}

\def\harry#1{{ #1}}
\def\yang#1{{ #1}}

\begin{document}

\twocolumn[
\icmltitle{Reinforcement Learning from Imperfect Demonstrations}



\icmlsetsymbol{equal}{*}

\begin{icmlauthorlist}
\icmlauthor{Yang Gao$^*$}{berkeley}
\icmlauthor{Huazhe(Harry) Xu$^*$}{berkeley}
\icmlauthor{Ji Lin}{tsinghua}
\icmlauthor{Fisher Yu}{berkeley}
\icmlauthor{Sergey Levine}{berkeley}
\icmlauthor{Trevor Darrell}{berkeley}
\end{icmlauthorlist}

\icmlaffiliation{berkeley}{Department of Electrical Engineering and Computer Science, UC Berkeley, CA, USA}
\icmlaffiliation{tsinghua}{Department of Electrical Engineering, Tsinghua University, Beijing, China}

\icmlcorrespondingauthor{Yang Gao}{yg@eecs.berkeley.edu}
\icmlcorrespondingauthor{Huazhe Xu}{huazhe\_xu@eecs.berkeley.edu}

\icmlkeywords{reinforcement learning, learning from demonstration}

\vskip 0.2in
]



\printAffiliationsAndNotice{\icmlEqualContribution} % otherwise use the standard text.

\begin{abstract}
Robust real-world  learning should benefit from both demonstrations and interactions with the environment. Current approaches to learning from demonstration and reward perform supervised learning on expert demonstration data and use reinforcement learning  to further improve performance based on the reward received from the environment. These tasks have divergent losses which are difficult to jointly optimize and such methods can be very sensitive to noisy demonstrations. We propose a unified reinforcement learning algorithm, Normalized Actor-Critic (NAC), that effectively normalizes the Q-function, reducing the Q-values of actions unseen in the demonstration data. NAC learns an initial policy network from demonstrations and refines the policy in the environment, surpassing the demonstrator's performance. Crucially, both learning from demonstration and interactive refinement use the same objective, unlike prior approaches that combine distinct supervised and reinforcement losses. This makes NAC robust to suboptimal demonstration data, since the method is not forced to mimic all of the examples in the dataset. We show that our unified reinforcement learning algorithm can learn robustly and outperform existing baselines when evaluated on several realistic driving games.

\end{abstract}

\section{Introduction}
Deep reinforcement learning (RL) has achieved significant success on many complex sequential decision-making problems. However, RL algorithms usually require a large amount of interactions with an environment to reach good performance \citep{kakade2003sample}; initial performance may be nearly random, clearly suboptimal, and often rather dangerous in real-world settings such as autonomous driving. Learning from demonstration is a well-known alternative, but typically does not leverage reward, and presumes relatively small-scale noise-free demonstrations. We develop a new robust algorithm that can learn value and policy functions from state, action and reward ($s, a, r$) signals that either come from imperfect demonstration data or the environment.

Recent efforts toward policy learning which does not suffer from a suboptimal initial performance generally leverage an initial phase of supervised learning and/or auxiliary task learning. Several previous efforts have shown that demonstrations can speed up RL by mimicking expert data with a temporal difference regularizer \citep{hester2017learning} or via gradient-free optimization \citep{pmlr-v78-ebrahimi17a}, yet these methods presume near-optimal demonstrations. \cite{jaderberg2016reinforcement} and \cite{shelhamer2016loss} obtained better initialization via auxiliary task losses (e.g., predicting environment dynamics) in a self-supervised manner; policy performance is still initially random with these approaches.

A simple combination of several distinct losses can learn from demonstrations; however, it is more appealing to have a single principled loss that is applicable to learning both from the demonstration and  from the environment. Our approach, Normalized Actor-Critic (NAC), uses a unified loss function to process both off-line demonstration data and on-line experience based on the underlying maximum entropy reinforcement learning framework \citep{toussaint2009robot,pmlr-v70-haarnoja17a, schulman2017equivalence}. Our approach enables robust learning from corrupted (or even partially adversarial) demonstrations that contains ($s, a, r$), because no assumption on the optimality of the data is required. A normalized formulation of the soft Q-learning gradient enables the NAC method, which can also be regarded as a variant of the policy gradient. 

We evaluate our approach in a toy Minecraft Game, as well as two realistic 3D simulated environments, Torcs and Grand Theft Auto V (GTA V), with either discrete states and tabular Q functions, or raw image input and  Q functions approximated by neural networks. Our experimental results  outperform previous approaches on driving tasks with only a modest amount of demonstrations while tolerating significant noise in the demonstrations, as our method utilizes rewards rather than simply imitates demonstrated behaviors. 

We summarize the contributions in this paper as follows.
\begin{itemize}
\item We propose the NAC method for learning from demonstrations and discover its practical advantages on a variety of environments.
\item To the best of our knowledge, we are the first to propose a unified objective, capable of learning from both demonstrations and environments, that outperforms methods including the ones with an explicit supervised imitation loss. 
\item Unlike other methods that utilize supervised learning to learn from demonstrations, our pure reinforcement learning method is robust to noisy demonstrations.
\end{itemize}

\section{Preliminaries}
In this section, we will briefly review the reinforcement learning techniques that our method is built on, including maximum entropy reinforcement learning and the soft Q-learning.
\subsection{Maximum Entropy Reinforcement Learning}
The reinforcement learning problem we consider is defined by a Markov decision process(MDP) \citep{thie1983markov}. Specifically, the MDP is characterized by a tuple $<\mathbb{S}$,$\mathbb{A}$,R,T,$\gamma>$, where $\mathbb{S}$ is the set of states, $\mathbb{A}$ is the set of actions, $R(s, a)$ is the reward function, $T(s, a, s') = P(s'|s, a)$ is the transition function and $\gamma$ is the reward discount factor. An agent interacts with the environment by taking an action at a given state, receiving the reward, and transiting to the next state. 

In the standard reinforcement learning setting \citep{sutton1998reinforcement}, the goal of an agent is to learn a  policy $\pi_{std}$, such that an agent maximizes the future discounted reward:
\begin{align}
\begin{gathered}
\pi_{std} = \mathop{\operatorname{argmax}}_{\pi} \sum_t \gamma^t \mathop{\mathbb{E}}_{s_t,a_t\sim\pi}[ R_t]
\end{gathered}
\end{align}
Maximum entropy policy learning \citep{ziebart2010modeling, pmlr-v70-haarnoja17a} uses an entropy augmented reward. The optimal policy will not only optimize for discounted future rewards, but also maximize the discounted future entropy of the action distribution:
\begin{align}
\begin{gathered}
 \pi_{ent} = \mathop{\operatorname{argmax}}_{\pi} \sum_t \gamma^t\mathop{\mathbb{E}}_{s_t,a_t\sim\pi}[R_t+\alpha H(\pi(\cdot|s_t))]
 \label{eq:soft_objective}
\end{gathered}
\end{align}
where $\alpha$ is a weighting term to balance the importance of the entropy. Unlike previous attempts that only adds the entropy term at a single time step, maximum entropy policy learning maximizes the discounted future entropy over the whole trajectory. Maximum entropy reinforcement learning has many benefits, such as better exploration in multi-modal problems and connections between Q-learning and the actor-critic method \citep{pmlr-v70-haarnoja17a,schulman2017equivalence}. 

\subsection{Soft Value Functions}
Since the maximum entropy RL paradigm augments the reward with an entropy term, the definition of the value functions naturally changes to
\begin{align}
Q_{\pi}(s,a)  &=R_0+ \mathop{\mathbb{E}}_{(s_t, a_t)\sim \pi}\sum_{t=1}^{\infty}\gamma^t (R_t+ \alpha H(\pi(\cdot|s_t))) \\
V_{\pi}(s) &=\mathop{\mathbb{E}}_{(s_t, a_t)\sim \pi}\sum_{t=0}^{\infty}\gamma^t (R_t+ \alpha H(\pi(\cdot|s_t))) 
\end{align}

where $\pi$ is some policy that value functions evaluate on. Given the state-action value function $Q^*(s, a)$ of the optimal policy,  \cite{ziebart2010modeling} shows that the optimal state value function and the optimal policy could be expressed as:
\begin{align}
V^*(s) &= \alpha \log \sum_{a}\exp(Q^*(s,a)/\alpha) \label{eq:optimal_v}\\
\pi^*(a|s) & = \exp((Q^*(s, a) - V^*(s))/\alpha) \label{eq:optimal_pi} 
\end{align}

\subsection{Soft Q-Learning and Policy Gradient}
\label{sec:soft_q_and_pg}
With the entropy augmented reward, one can derive the soft versions of Q-learning \citep{haarnoja2017reinforcement} and policy gradient. The soft Q-learning gradient is given by
\begin{align}
\label{eq:soft_q}
\nabla_\theta Q_{\theta}(s,a) (Q_{\theta}(s,a) - \hat{Q}(s,a))
\end{align}
where $\hat{Q}(s, a)$ is a bootstrapped Q-value estimate obtained by $R(s, a) + \gamma V_Q(s')$. Here, $R(s, a)$ is the reward received from the environment, $V_Q$ is computed from $Q_\theta(s, a)$ with Equation~ \eqref{eq:optimal_v}. We can also derive a policy gradient, which includes the gradient of form:
\begin{align}
\mathbb{E} \left [ \sum_{t=0}^\infty \nabla_\theta \log \pi_\theta(\hat{Q}_\pi - b(s_t)) + \alpha \nabla_\theta H(\pi_\theta(\cdot | s_t)) \right ] \label{eq:soft_pg}
\end{align}
where $b(s_t)$ is some arbitrary baseline \citep{schulman2017equivalence}.

\section{Robust Learning from Demonstration and Reward}
\label{sec:our_method}
\begin{algorithm}[t]
   \caption{Normalized Actor-Critic for Learning from Demonstration}
   \begin{algorithmic}
    \STATE\textbf{$\theta$: parameters for the rapid Q network, $\theta'$: parameters for the target Q network, $\mathcal{D}$: demonstrations collected by human or a trained policy network, $T$: target network update frequency, $\mathcal{M}$: replay buffer, $k$: number of steps to train on the demonstrations}
    \FOR{step t $\in$ \{1, 2, ...\}}
        	\IF {t $\leq$ $k$}
            	\STATE Sample a mini-batch of transitions from $\mathcal{D}$
            \ELSE 
            	\STATE Start from s, sample $a$ from $\pi$, execute $a$, observe ($s'$,$r$) and store $(s,a,r,s')$ in $\mathcal{M}$
            	\STATE Sample a mini-batch of transitions from $\mathcal{M}$ 
            \ENDIF 
            \STATE {Update $\theta$ with gradient: $\nabla_\theta J_{PG}+\nabla_\theta J_V$}
            \IF {t mod T = 0} 
           			\STATE $\theta' $ $\leftarrow$ $\theta$
            \ENDIF
     \ENDFOR
     \end{algorithmic}
     \label{alg:our_method}     
\end{algorithm}

Given a set of demonstrations that contains ($s, a, r, s'$) and the corresponding environment, an agent should perform appropriate actions when it starts the interaction, and continue to improve. Although a number of off-policy RL algorithms could in principle be used to learn directly from off-policy demonstration data, standard methods can suffer from extremely poor performance when trained entirely on demonstration data. This can happen when the demonstration set is a strongly biased sample of the environment transitions, which violates the assumptions of many off-policy RL methods. Although they are closely related, off-policy learning and learning from demonstrations are different problems. In section \ref{results}, we show that Q-learning completely fails on the demonstration data. The intuition behind this problem is that if the Q-function is trained only on good data, it has no way to understand why the action taken is appropriate: it will assign a high Q-value, but will not necessarily assign a low Q-value to other alternative actions.

The framework of soft optimality provides us with a natural mechanism to mitigate this problem by \emph{normalizing} the Q-function over the actions. Our approach, Normalized Actor-Critic (NAC), utilizes soft policy gradient formulations described in Section \ref{sec:soft_q_and_pg} to obtain a Q-function gradient that reduces the Q-values of actions that were not observed along the demonstrations. In other words, without data to indicate otherwise, NAC will opt to follow the demonstrations. This method is a well-defined RL algorithm without any auxiliary supervised loss and hence, it is able to learn without bias in the face of low-quality demonstration data. We will first describe our algorithm, and then discuss why it performs well when trained on the demonstrations.

\subsection{Normalized Actor-Critic for Learning from Demonstration}
\label{actor_critic}
We propose a unified learning from demonstration approach, which applies the normalized actor-critic updates to both off policy demonstrations and in-environment transitions. The NAC method is derived from the soft policy gradient objective with a $Q$ function parametrization. Specifically, we take gradient steps to maximize the future reward objective (Equation \eqref{eq:soft_objective}), and parametrize $\pi$ and $V$ in terms of $Q$ (Equation \eqref{eq:optimal_pi} and \eqref{eq:optimal_v}). As derived in the appendix, the updates for the actor and critic are:
\begin{align}
&\nabla_\theta J_{PG}=  \mathop{\mathbb{E}}_{s, a \sim \pi_Q} \left [ (\nabla_\theta Q(s, a) - \nabla_\theta V_Q(s)) ( Q(s,a) - \hat{Q}) \right ] \label{eq:our_pg} \\
&\nabla_\theta J_V = \mathop{\mathbb{E}}_s \left [ \nabla_\theta \frac{1}{2} (V_Q(s) - \hat{V}(s))^2  \right ] \label{eq:our_v} 
\end{align}
where $V_Q$ and $\pi_Q$ are deterministic functions of $Q$: 
\begin{align}
V_Q(s) &= \alpha \log \sum_a \exp(Q(s, a)/\alpha) \\
\pi_Q(a|s) &=~\exp((Q(s,a) - V_Q(s))/\alpha)  
\end{align}

$\hat{Q}(s,a), \hat{V}(s)$ are obtained by:
\begin{align}
&\hat{Q}(s, a) = R(s, a) + \gamma V_Q(s') \\
&\hat{V}(s) = \mathop{\mathbb{E}}_{a\sim \pi_Q} \left [ R(s, a) +\gamma V_Q(s') \right ] +  \alpha H(\pi_Q(\cdot|s))
\end{align}

\label{sec:rlfd_weighting}
The difference is the $\nabla_\theta V(s)$ term comparing NAC's actor update term (Equation \eqref{eq:our_pg}) with the soft Q update (Equation \eqref{eq:soft_q}). We emphasize the normalization effect of this term: it avoids pushing up the $Q$-values of actions that are not demonstrated. The mechanism is explained  in Section \ref{sec: analysis}.

The expectations in Eq. \eqref{eq:our_pg} and Eq. \eqref{eq:our_v} are taken with respect to $\pi_Q$. In the demonstration set, we only have transition samples from the behavioral policy $\mu(a|s)$. To have a proper policy gradient algorithm, we can employ importance sampling to correct the mismatch. To be specific, when estimating $\mathbb{E}_{(s, a)\sim \pi_Q} \left [ f(s, a) \right ]$, we estimate $\mathbb{E}_{(s, a)\sim \mu} \left [ f(s, a) \beta  \right ]$ , where $\beta =  \min \left \{ \frac{\pi_Q(a|s)}{\mu(a|s)}, c \right \}$ and $c$ is some constant that prevents the importance ratio from being too large. Although the importance weights are needed to formalize our method as a proper policy gradient algorithm, we find in our empirical evaluation that the inclusion of these weights consistently reduces the performance of our method. We found that omitting the weights results in better final performance even when training entirely on demonstration data. For this reason, our final algorithm does not use importance sampling.

We summarize the proposed method in Algorithm~\ref{alg:our_method}. Our method uses samples from the demonstrations and the replay buffer, rather than restricting the samples to be on policy as in standard actor-critic methods. Similar to DQN, we utilize a target network to compute $\hat{Q}(s, a)$ and $\hat{V}(s)$, which stabilizes the training process. 

\subsection{Analysis of the Method  }
\label{sec: analysis}
We provide an intuitive analysis in this section to explain why our method can learn from demonstrations while other reinforcement learning methods, such as Q-learning cannot.
The states and actions in the demonstrations generally have higher Q-values than other states.
Q-learning will push up $Q(s, a)$ values in a sampled state $s$. However, if the values for the bad actions are not observed, the Q-function has no way of knowing whether the \emph{action} itself is good, or whether all actions in that \emph{state} are good, so the demonstrated action will not necessarily have a higher Q-value than other actions in the demonstrated state.

Comparing the actor update (Eq.~\eqref{eq:our_pg}) of our method with the soft Q-learning (Eq. \eqref{eq:soft_q}) update, our method includes an extra term in the gradient: $-\nabla_\theta V_Q(s)$.
This term falls out naturally when we derive the update from a policy gradient algorithm, rather than a Q-learning algorithm.
Intuitively, this term will decrease $V_Q(s)$ when increasing $Q(s, a)$ and vice versa, since $\nabla_\theta Q(s,a)$ and $-\nabla_\theta V_Q(s)$ have different signs. Because $V_Q(s) = \alpha \log \sum_a \exp(Q(s, a)/\alpha)$, decreasing $V_Q(s)$ will prevent $Q(s,a)$ from increasing for the actions that are not in the demonstrations. That is why the aforementioned normalization effect emerges with the extra $\nabla_\theta V_Q(s)$ term. 

Besides having the normalizing behaviors, NAC is also less sensitive to noisy demonstrations. Since NAC is an RL algorithm, it is naturally resistant to poor behaviors. One could also see this with similar analysis as above. When there is a negative reward in the demonstrations, $Q(s, a)$ tends to decrease and $V_Q(s)$ tends to increase, hence having the normalizing behavior in a reverse direction. 

Moreover, NAC provides a single and principled approach to learn from both demonstrations and environments. It avoids the use of imitation learning. Therefore, besides its natural robustness to imperfect demonstrations, it is a more unified approach comparing with other methods.

\begin{figure}[t]
    \centering
    \begin{subfigure}[t]{0.4\textwidth}
        \includegraphics[width=\textwidth]{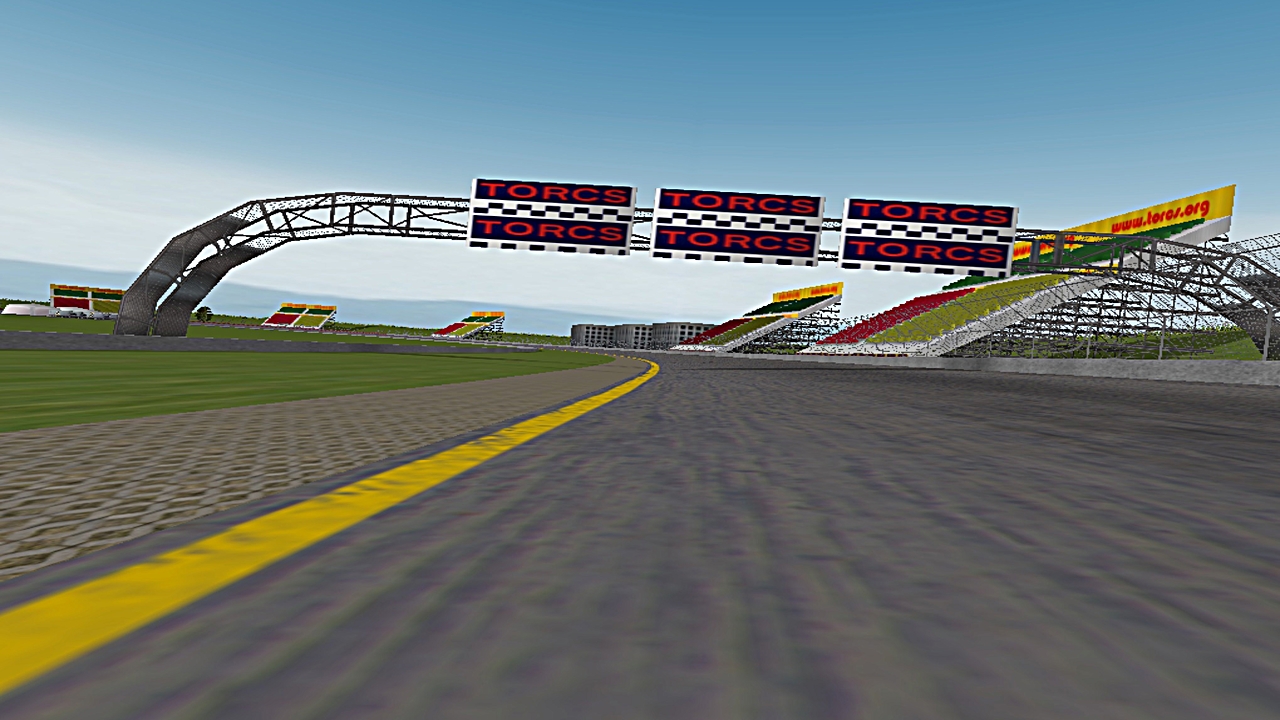}
    \end{subfigure}
    
    \begin{subfigure}[t]{0.4\textwidth}
        \includegraphics[width=\textwidth]{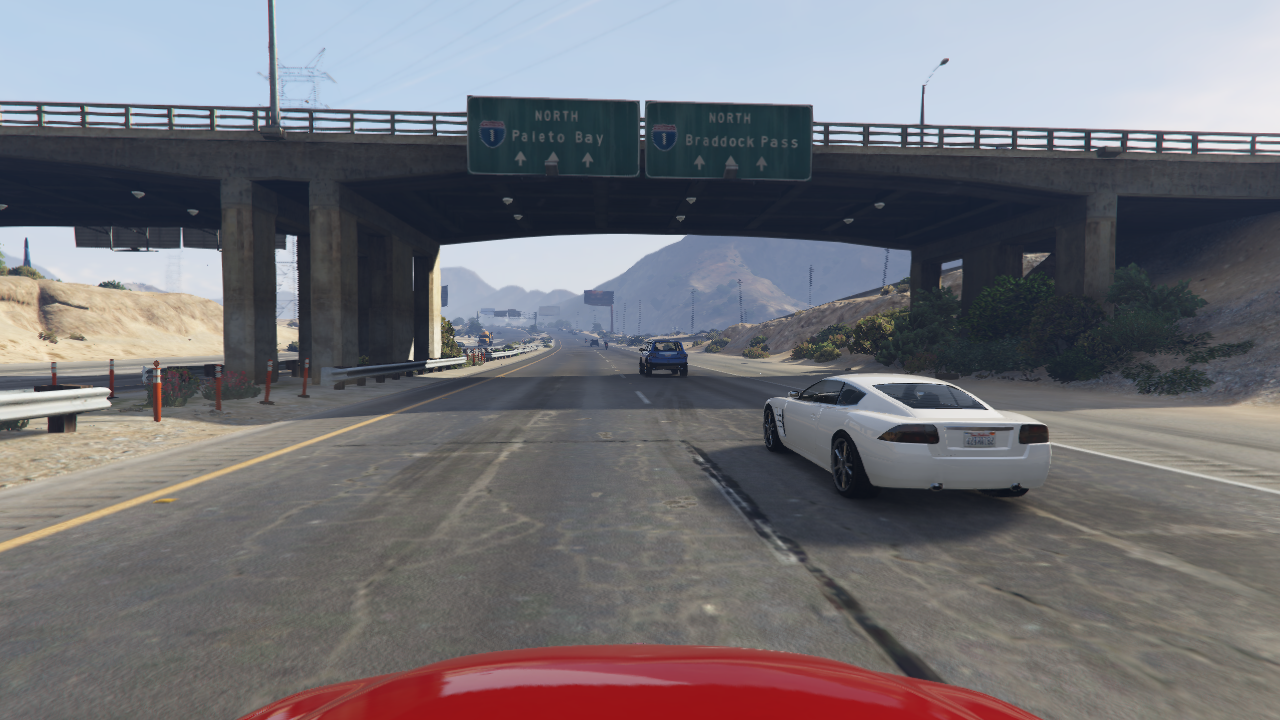}
    \end{subfigure}
   
    \caption{Sample frames from Torcs (upper) and GTA (lower).}
    \label{fig:torcsgta}
    \vskip -.4cm
\end{figure}

\begin{figure}[t]
    \centering
    \includegraphics[width=0.4\textwidth]{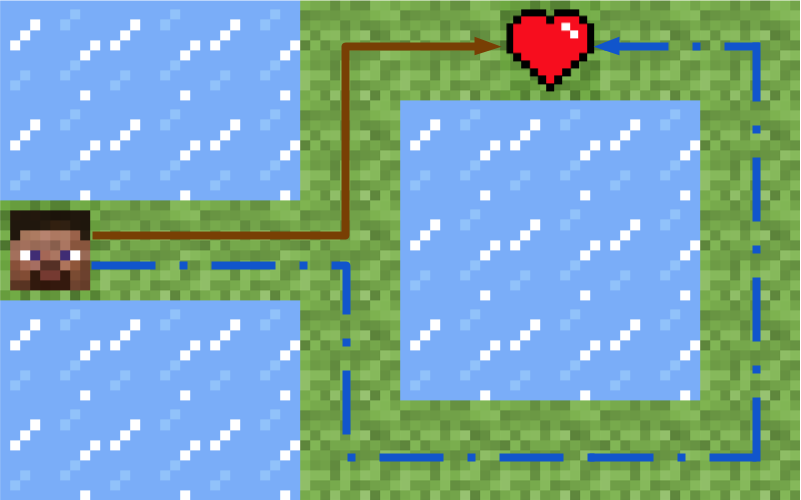}
    \caption{The Toy Minecraft environment. We aim to learn a policy that moves the agent to the goal. Two possible paths are shown, the shorter optimal one (solid, brown) and the longer sub-optimal one (dashed, blue). See text for details about the environment (Sec. \ref{sec:toy_minecraft}) and comparison between NAC and DQfD on this environment (Sec. \ref{sec:exp_minecraft}).}
    \label{fig:minecraft}
    \vskip -.7cm
\end{figure}

\section{Related Work}
\subsection{Maximum entropy Reinforcement Learning}
Maximum entropy reinforcement learning has been explored in a number of prior works~\cite{todorov2008general,toussaint2009robot,ziebart2008maximum}, including several recent works that extend it into a deep reinforcement learning setting~\cite{nachum2017bridging,pmlr-v70-haarnoja17a,schulman2017equivalence}. However, most of those works do not deal with the learning from demonstration settings. \cite{pmlr-v70-haarnoja17a} and \cite{schulman2017equivalence} propose maximum entropy RL methods to learn from environments. PCL \citep{nachum2017bridging} is the only prior work that studies the learning from demonstration task with the maximum entropy RL framework. Unlike their method where the loss is derived from an objective similar to Bellman error, our method is derived from policy gradient. Instead of minimizing Bellman errors, policy gradient directly optimizes future accumulated reward. As shown in Section \ref{results}, our method has large performance advantage compared with PCL, due to the different objective function. 

Our method not only admits a unified objective on both demonstrations and environments but also performs better than alternative methods, such as PCL \citep{nachum2017bridging} and DQfD \citep{hester2017learning}. To the best of our knowledge, our proposed method is the first unified method across demonstrations and environments that outperforms methods including the ones with explicit supervised imitation loss  such as DQfD.

\subsection{Learning from Demonstration}
Most of the prior learning from demonstration efforts assume the demonstrations are perfect, \textit{i.e.} the ultimate goal is to copy the behaviors from the demonstrations. Imitation learning is one of such approaches, examples including ~\cite{xu2016end,bojarski2016end}. Extensions such as DAGGER \citep{ross2011reduction} are proposed to have the expert in the loop, which further improves the agent's performance. Recently, \cite{ho2016generative,wang2017robust,ziebart2008maximum} explore an adversarial paradigm for the behavior cloning method. Another popular paradigm is Inverse Reinforcement Learning (IRL) \citep{ng2000algorithms,abbeel2004apprenticeship,ziebart2008maximum}. IRL learns a reward model which explains the demonstrations as optimal behavior.

Instead of assuming that the demonstrations are perfect, our pure RL method allows imperfect demonstrations. Our method \emph{learns} which part of the demonstrations is good and which part is bad, unlike the methods that simply imitate the demonstrated behaviors. We follow the Reinforcement Learning with Expert Demonstrations (RLED) framework~\citep{chemali2015direct,kim2013learning,piot2014boosted}, where both rewards and actions are available in the demonstrations. The extra reward in the demonstrations allows our method to be aware of poor behaviors in the demonstrations. DQfD \citep{hester2017learning} is a recent method that also uses rewards in the demonstrations. It combines an imitation hinge loss with the Q-learning loss in order to learn from demonstrations and transfer to environments smoothly. Due to the use of the imitation loss, DQfD is more sensitive to noisy demonstrations, as we show in the experiment section.   

\subsection{Off-policy Learning}
It is tempting to apply various off-policy methods to the problem of learning from demonstration, such as policy gradient variants \citep{gu2017interpolated,gu2016q,degris2012off,wang2016sample}, Q-learning \citep{watkins1992q} and Retrace \citep{munos2016safe}. However, we emphasize that off-policy learning and learning from demonstration are different problems. For most of the off-policy methods, their convergence relies on the assumption of visiting each $(s, a)$ pair infinitely many times. In the learning from demonstration setting, the samples are highly biased and off-policy method can fail to learn anything from the demonstrations, as we explained the Q-learning case in Section \ref{sec:our_method}.

\section{Results}
\label{results}
Our experiments address several  questions: (1) Can NAC benefit from both demonstrations and rewards? (2) Is NAC robust to ill-behaved demonstrations? (3) Can NAC learn meaningful behaviors with a limited amount of demonstrations? We compare our algorithm with DQfD \citep{hester2017learning}, which has been shown to learn efficiently from demonstrations and to preserve performance while acting in an environment. Other methods include supervised behavioral cloning method, Q-learning, soft Q-learning, the version of our method with importance sampling weighting, PCL and Q-learning, soft Q-learning without demonstrations.

\subsection{Environments}
We evaluate our result in a grid world, the toy Minecraft (Fig. \ref{fig:minecraft}), as well as two realistic 3D simulated environments, Torcs and Grand Theft Auto V (GTA V) shown in Figure \ref{fig:torcsgta}. 

\label{sec:toy_minecraft}
\textbf{Toy Minecraft}: The toy Minecraft is a customized grid world environment. As shown in Figure \ref{fig:minecraft}, the agent starts from the left and would like to reach the final goal (marked as a heart). The agent can walk on the green grass and go into the blue water ends the episode.  The input to the agent is its current $(x, y)$ location. At each step, the agent can move \emph{Up}, \emph{Down}, \emph{Left} or \emph{Right}. It gets a reward of $1$ when reaching the goal, $0$ otherwise. For more details, please refer to the OpenAI gym Frozen-Lake environment \citep{brockman2016openai}.

\textbf{Torcs}: Torcs is an open-source racing game that has been used widely as an experimental environment for driving. The goal of the agent is to drive as fast as possible on the track while avoiding crashes. We use an oval two-lane racing venue in our experiments. The input to the agent is an 84$\times$84 gray scale image. The agent controls the vehicle at 5Hz, and at each step, it chooses from a set of 9 actions which is a Cartesian product between \{left, no-op, right\} and \{up, no-op, down\}. We design a dense driving reward function that encourages the car to follow the lane and to avoid collision with obstacles. 
\footnote{$reward = (1-\mathbbm{1}_{damage})[(\cos\theta-\sin\theta-lane\_ratio)$ $\times speed ] + \mathbbm{1}_{damage} \left [-10 \right ]$, where $\mathbbm{1}_{damage}$ is an indicator function of whether the vehicle is damaged at the current state. $lane\_ratio$ is the ratio between distance to lane center and lane width. $\theta$ is the angle between the vehicle heading direction and the road direction.}  

\textbf{GTA}: Grand Theft Auto is an action-adventure video game with goals similar in part to the Torcs game, but with a more diverse and realistic surrounding environment, including the presence of other vehicles, buildings, and bridges. The agent observes 84$\times$84 RGB images from the environment. It chooses from 7 possible actions from \{left-up, up, right-up, left, no-op, right, down\} at 6Hz. We use the same reward function as in Torcs.

\begin{figure*}[t]
    \centering
    \begin{subfigure}[t]{0.5\textwidth}
        \centering
        \includegraphics[width=\textwidth]{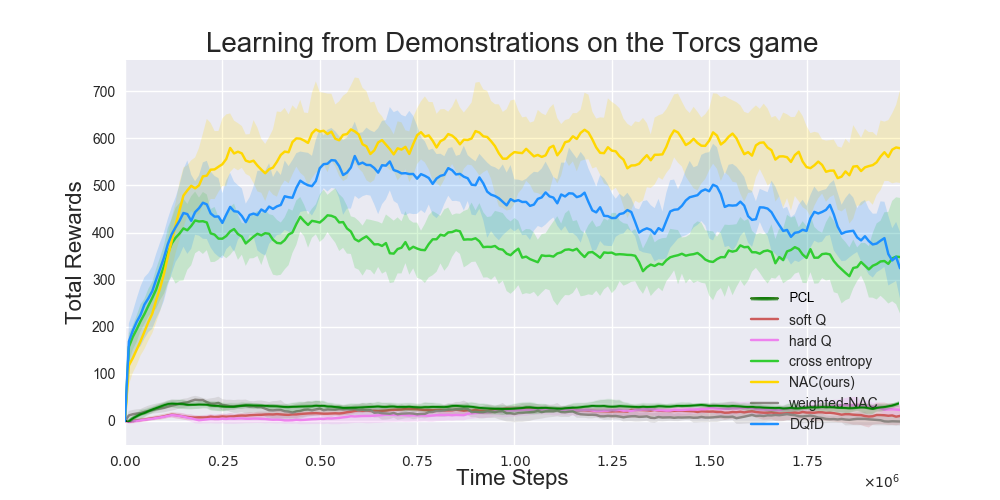}
    \end{subfigure}%
    ~ 
    \begin{subfigure}[t]{0.5\textwidth}
        \centering
        \includegraphics[width=\textwidth]{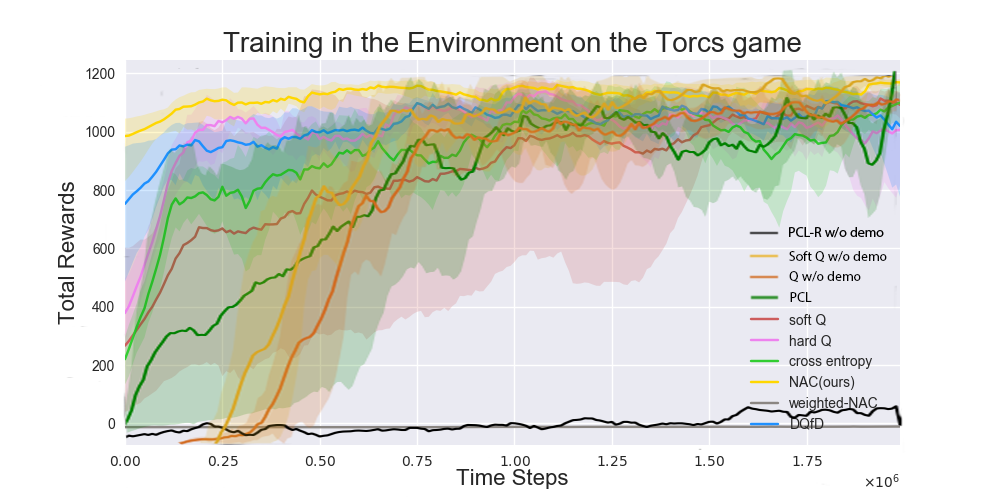}
    \end{subfigure}
    \caption{Performances on the Torcs game. The x-axis shows the
training iterations. The y-axis shows the average total rewards. Solid lines are average values over 10 random seeds. Shaded regions correspond to one standard deviation. The left figure shows the performance for each agent when they only learn from demonstrations, while the right one shows the performance for each agent when they interact with the environments after learning from demonstrations. \yang{Our method consistently outperforms other methods in both cases. } }
    \label{fig:rlfd_baselines}
\end{figure*}

\subsection{Comparisons}
We compare our approach with the following methods:
\begin{itemize}[leftmargin=*]
\item \textbf{DQfD:} the method proposed by \cite{hester2017learning}. For the learning from demonstration phase, DQfD combines a hinge loss with a temporal difference (TD) loss. For the finetuning-in-environment phase, DQfD combines a hinge loss on demonstrations and a TD loss on both the demonstrations and the policy-generated data. To alleviate over-fitting issues, we also include weight decay following the original paper.

\item \textbf{Q-learning:} the classic DQN method \citep{mnih2015human}. We first train DQN with the demonstrations in a replay buffer and then finetune in the environment with regular Q-learning. Similar to DQfD, we use a constant exploration ratio of 0.01 in the finetuning phase to preserve the performance obtained from the demonstrations. \harry{We also train from scratch a baseline DQN in the environment, without any demonstration. } 

\item \textbf{Soft Q-learning:} similar to the Q-learning method, but with an entropy regularized reward. This is the method proposed by \cite{haarnoja2017reinforcement,schulman2017equivalence}. \harry{ We also include the soft Q-learning trained without demonstration, as another baseline. }

\item \textbf{Behavior cloning with Q-learning:} the naive way of combining cross-entropy loss with Q-learning. First we perform behavior cloning with cross-entropy loss on the demonstrations. Then we treat the logit activations prior the softmax layer as an initialization of the Q function and finetune with  regular Q-learning in the environment. 

\item \textbf{Normalized actor-critic with importance sampling:} the NAC method with the importance sampling weighting term mentioned in Sec \ref{sec:rlfd_weighting}. The importance weighting term is used to correct the action distribution mismatch between the demonstration and the current policy. 

\harry{\item \textbf{Path Consistency Learning (PCL):} the PCL \citep{nachum2017bridging} method that minimizes the soft path consistency loss. The method proposed in the original paper (denoted as \emph{PCL-R}) does not utilize a target network. We find that PCL-R does not work when it is trained from scratch in the visually complex environment. We stabilize it by adding a target network (denoted as \emph{PCL}), similar to \cite{haarnoja2017reinforcement}. }
\end{itemize}

\subsection{Experiments on Toy Minecraft}
\label{sec:exp_minecraft}

To understand the basic properties of our proposed method, we design the toy Minecraft environment. In this experiment, the state is simply the location of the agent. We use a tabular $Q$ function. With those settings, we hope to reveal some differences between our NAC algorithm and algorithms that incorporate supervised loss.

As shown in Figure \ref{fig:minecraft}, there are only two paths to reach the goal. In terms of the discounted reward, the shorter path is more favorable. To make the problem more interesting, we provide the longer suboptimal path as the demonstrations. We found that in the learning from demonstration phase, both DQfD and NAC have learned the suboptimal path since both methods do not have access to the environment and could not possibly figure out the optimal path. When the two methods finetune their policies in the environment, NAC succeeds in finding the optimal path, while DQfD stucks with the suboptimal one. It is because DQfD has the imitation loss, thus preventing it from deviating from the original solution.

\begin{figure*}[t]
  \centering
  \begin{subfigure}[t]{0.48\textwidth}
  \centering
  \includegraphics[width=\textwidth]{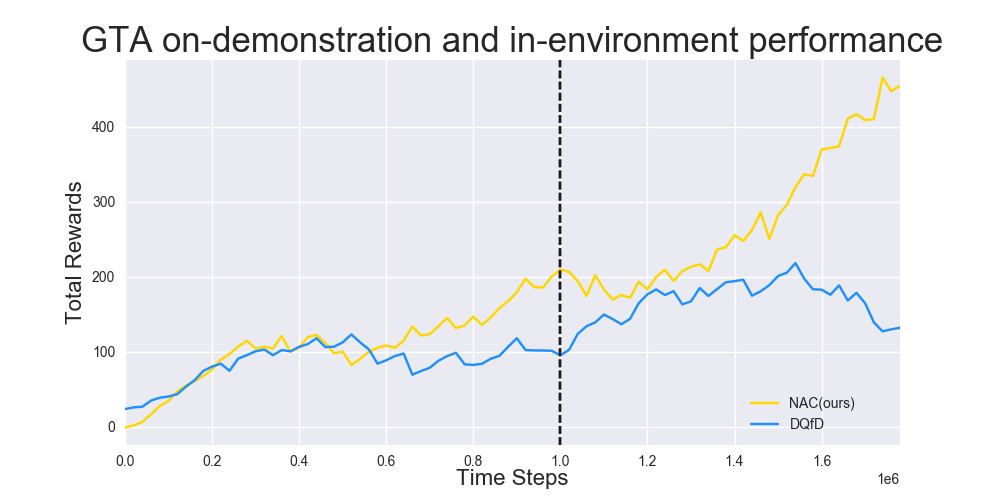}
  \caption{\yang{The on-demonstration and in-environment performance of the NAC and DQfD methods on GTA. The vertical line separates the learning from demonstration phase and finetuning in environment phase. Our method consistently outperforms DQfD in both phases.}}
  \label{fig:GTA}
  \end{subfigure}
  ~
   \begin{subfigure}[t]{0.48\textwidth}
    \centering
    \includegraphics[width=\textwidth]{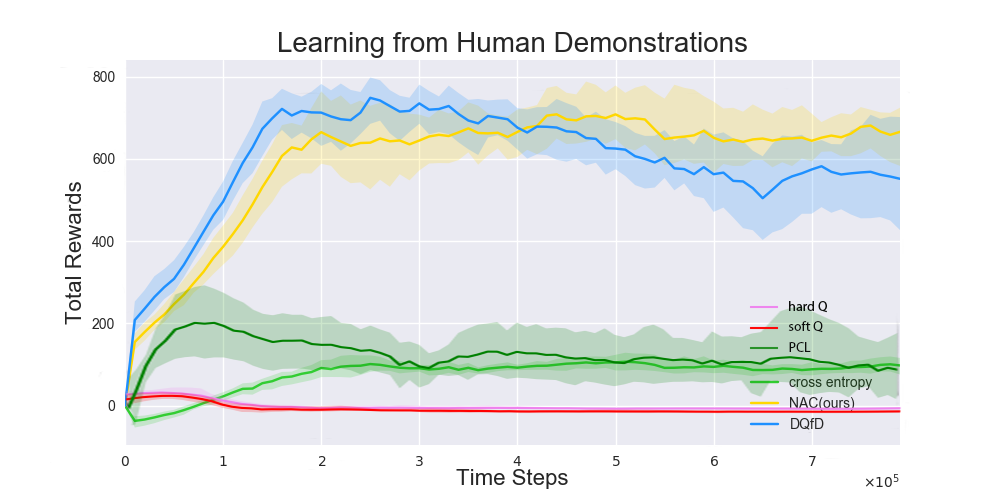}
    \caption{Performances on the Torcs game with human demonstrations. \yang{DQfD performs well in the beginning, but overfits in the end. The behavior cloning method is much worse than NAC and DQfD. Our NAC method performs best at convergence. }}
    \label{fig:human_baselines}
    \end{subfigure}
    \caption{Performance on GTA (left) and performance on Torcs with human demonstrations (right)}
\end{figure*}

\subsection{Comparison to Other Methods}
\label{sec:compare_baseline}
We compare our NAC method with other methods on 300k transitions. The demonstrations are collected by a trained Q-learning expert policy. We execute the policy in the environment to collect demonstrations. To avoid deterministic executions of the expert policy, we sample an action randomly with probability 0.01. 

To explicitly compare different methods, we show separate figures for performances on the demonstrations and inside the environments. In Fig \ref{fig:rlfd_baselines}, we show that our method performs better than other methods on demonstrations. When we start finetuning, the performance of our method continues to increase and reaches peak performance faster than other methods. DQfD~\citep{hester2017learning} has similar behavior to ours but has lower performance. Behavior cloning learns well on demonstrations, but it has a significant performance drop while interacting with environments. All the methods can ultimately learn by interacting with the environment but only our method and DQfD start from a relatively high performance. Empirically, we found that the importance weighted NAC method does not perform as well as NAC. The reason might be the decrease in the gradient bias is not offset sufficiently by the increase in the gradient variance. Without the demonstration data, Q-learning (Q w/o demo) and soft Q-learning (soft-Q w/o demo) suffer from low performance during the initial interactions with the environment. The original PCL-R method (PCL-R w/o demo) fails to learn even when trained from scratch in the environments. The improved PCL method (PCL) is not able to learn on the demonstrations, but it can learn in the environment. 

\yang{We also test our method on the challenging GTA environment, where both the visual input and the game logic are more complex. Due to the limit of the environment execution speed, we only compare our method with DQfD. As shown in Fig. \ref{fig:GTA}, our method outperforms DQfD both on the demonstrations and inside the environment.}

\subsection{Learning from Human Demonstrations}
For many practical problems, such as autonomous driving, we might have a large number of human demonstrations, 
but no demonstration available from a trained agent at all. In contrast to a scripted agent, humans usually perform actions diversely, both from multiple individuals (e.g. conservative players will slow down before a U-turn; aggressive players will not) and a single individual (e.g. a player may randomly turn or go straight at an intersection). Many learning from demonstration methods do not study this challenging case, such as \cite{ho2016generative}. We study how different methods perform with diverse demonstrations. To collect human demonstrations, we asked 3 non-expert human players to play TORCS for 3 hours each. Human players control the game with the combination of four arrow keys, at 5Hz, the same rate as the trained agent. In total, we collected around 150k transitions. \yang{Among them, 4.5k transitions are used as a validation set to monitor the Bellman error. } Comparing with data collected from a trained agent, the data is more diverse and the quality of the demonstrations improves naturally when the players get familiar with the game.  

In Fig. \ref{fig:human_baselines}, we observe that the behavior cloning method performs much worse than NAC and DQfD. DQfD initially is better than our method but later is surpassed by the NAC method quickly, which might be caused by the supervised hinge loss being harmful when demonstrations are suboptimal. Similar to the policy generated demonstrations case, PCL, hard Q-learning and soft Q-learning do not perform well. 

\subsection{Effects of Imperfect Demonstrations}
\begin{figure*}[t]
    \centering
    \begin{subfigure}[t]{0.5\textwidth}
        \centering
        \includegraphics[width=\textwidth]{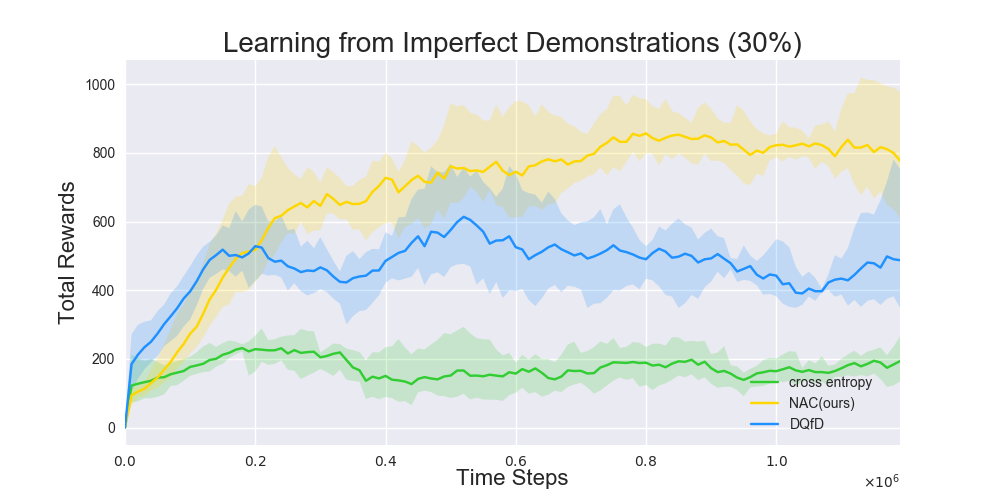}
    \end{subfigure}%
    ~
    \begin{subfigure}[t]{0.5\textwidth}
        \centering
        \includegraphics[width=\textwidth]{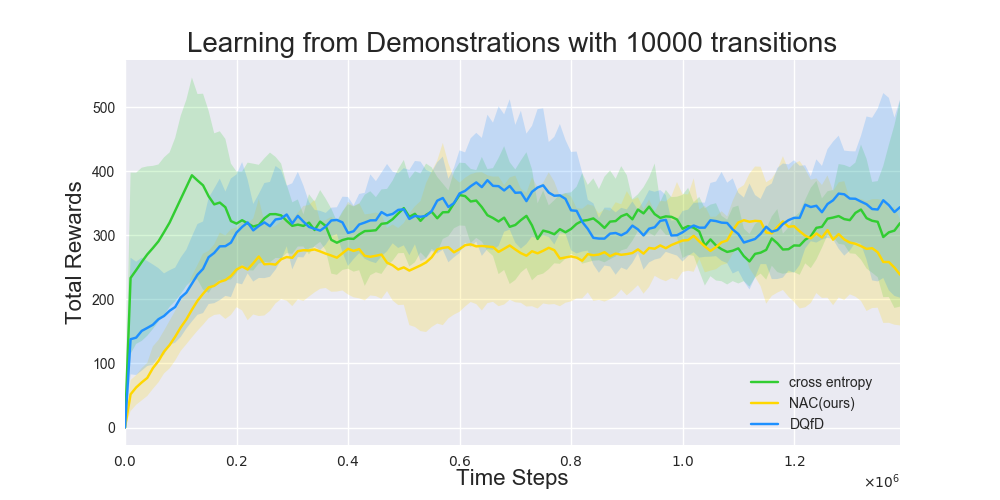}
    \end{subfigure}
    \caption{\yang{Left: Learning from imperfect data when the imperfectness is 30\%. Our NAC method does not clone suboptimal behaviors and thus outperforms DQfD and behavior cloning. Right: Learning from a limit amount of demonstrations. Even with only 30 minutes (10k transitions) of experience, our method could still learn a policy that is comparable with supervised learning methods. More results are available in the appendix, including 50\% and 80\% imperfect data ablations, as well as 150k and 300k data amount studies. }}
    \label{fig:imperfect_and_amount}
\end{figure*}
\label{sec: imperfect}

In the real world,  collected demonstrations might be far from optimal. The human demonstrations above have already shown that imperfect demonstrations could have a large effect on performance. To study this phenomenon in a principled manner, we collect a few versions of demonstrations with varying degrees of noise. When collecting the demonstrations with the trained Q agent, we corrupt a certain percentage of the demonstrations by choosing non-optimal actions ($\mathop{argmin}_a Q(s,a)$). \harry{The data corruption process is conducted while interacting with the environment; therefore, the error will affect the collection of the following steps.} We get 3 sets of $\{30\%, 50\%, 80\%\}$ percentage of imperfect data. In the left of Fig. \ref{fig:imperfect_and_amount}, we show that our method performs well compared with DQfD and behavior cloning methods. Supervised behavior cloning method is heavily influenced by the imperfect demonstrations. DQfD is also heavily affected, but not as severely as the behavior cloning. NAC is robust because it does not imitate the suboptimal behaviors. The results for 50\% and 80\% percentage of imperfect data are similar, and they are available in the appendix. 

\subsection{Effects of Demonstration Size}
In this section, we show comparisons among our method and other methods with different amounts of demonstration data. We use a trained agent to collect three sets of demonstrations which include 10k, 150k, and 300k transitions each.  In the experiments, we find that our algorithm performs well when the amount of data is large and is comparable to supervised methods even with a limited amount of data. In Fig. \ref{fig:imperfect_and_amount} (right), we show when there are extremely limited amounts of demonstration data (10k transitions or 30 minutes of experience), our method performs on par with supervised methods. In the appendix, we show the results for 150k and 300k transitions: our method outperforms the baselines by a large margin with 300k transitions. In summary, our method can learn from small amounts of demonstration data and dominates in terms of performance when there is a sufficient amount of data.

\subsection{Effects of Reward Choice}
\begin{figure}[t]
    \centering
    \includegraphics[width=0.45\textwidth, height=5cm]{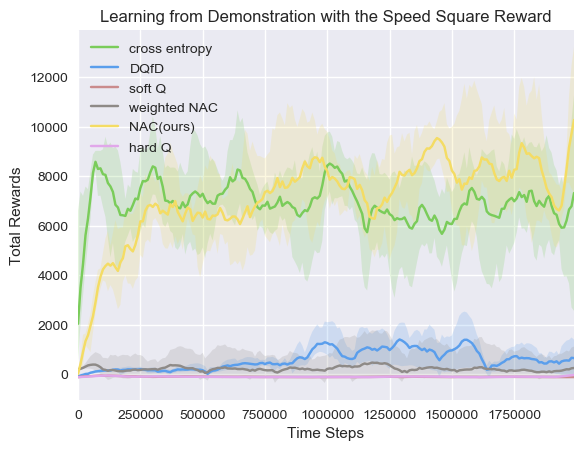}    
    \caption{Similar to Figure \ref{fig:rlfd_baselines} (left), we compare NAC with other methods when only learning from the demonstrations, except that we use a different reward: $speed^2$. Our method still performs the best. }    
    \label{fig:reward_choice}
    \vskip -.5cm
\end{figure}

In the above experiments, we adopt a natural reward: it maximizes speed along the lane, minimizes speed perpendicular to the lane and penalizes when the agent hits anything. However, very informative rewards are not available under many conditions. In this section, we study whether our method is robust to a less informative reward. We change the reward function to be the square of the speed of an agent, irrespective of the speed's direction. This reward encourages the agent to drive fast, however, it is difficult to work with because the agent has to learn by itself that driving off-road or hitting obstacles reduce its future speed. It is also hard because $speed^2$ has a large numerical range. Figure \ref{fig:reward_choice} shows that NAC method still performs the best at convergence, while DQfD suffers from severe performance degeneration.

\section{Conclusion}
We proposed a Normalized Actor-Critic algorithm for reinforcement learning from demonstrations. Our algorithm provides a unified approach for learning from reward and demonstrations, and is robust to  potentially suboptimal demonstration data. An agent can be fine-tuned with rewards after training on demonstrations by simply continuing to perform the same algorithm on on-policy data. Our algorithm preserves and improves the behaviors learned from demonstrations while receiving reward through  interaction with an environment.

\bibliography{example_paper}
\bibliographystyle{icml2018}
\clearpage

\section{Appendix}
\subsection{Normalized Actor-Critic with Q Parametrization}
\label{app:q_param}
Usually the actor-critic method parametrizes $\pi(s, a)$ and $V(s)$ with a neural network that has two heads. In this section, we explore an alternative parametrization: Q-parametrization. Instead of outputting $\pi$ and $V$ directly, the neural network computes $Q(s, a)$. We parametrize $\pi$ and $V$ based on $Q$ by specifying a fixed mathematical transform:

\begin{align}
V_Q(s) &= \alpha \log \sum_a \exp(Q(s, a)/\alpha)\label{eq:13} \\ 
\pi_Q(a|s) & = \exp((Q(s,a) - V_Q(s))/\alpha) \label{eq:14}
\end{align}

Note that the Q-parametrization we propose here can be seen as a specific design of the  network architecture. Instead of allowing the  net to output arbitrary $\pi(s, a)$ and $V(s)$ values, we restrict the network to only output $\pi(s, a)$ and $V(s)$ pairs that satisfy the above relationship. This extra restriction will not harm the network's ability to learn since the above relationship has to be satisfied at the optimal solution\citep{schulman2017equivalence,pmlr-v70-haarnoja17a,nachum2017bridging}.

Based on the Q-parametrization, we can derive the update of the actor. Note that we assume the behavioral policy is $\pi_Q$, and we sample one step out of a trajectory, thus dropping the subscript $t$. The goal is to maximize expected future reward, thus taking gradient of it we get:

\begin{align}
\begin{aligned}
& \nabla \mathbb{E}_{s, a  \sim \pi_Q} \left [ R(s,a) \right ]  \\
= & \mathbb{E}_{s,a \sim  \pi_Q} \left [ R(s,a) \nabla \log_\theta p(a, s|\pi_Q) \right ]\\
\approx & \mathbb{E}_{s,a \sim  \pi_Q} \left [ R(s,a) \nabla_\theta \log \pi_Q(a|s) \right ]
\end{aligned}
\end{align}

where the last step ignores the state distribution, thus an approximation. By adding some baseline functions, it turns to the following format, where $\hat{Q}(s, a) = R(s, a) + \gamma V_Q(s')$:
\begin{align}
\begin{aligned}
&\mathbb{E}_{s,a} \left[ \nabla_\theta \log \pi_Q(a|s) (\hat{Q}(s,a) - b(s)) \right]  \\
=& \mathbb{E}_{s} \left [ \sum_a \pi_Q(a|s) \nabla_\theta \log \pi_Q(a|s) (\hat{Q}(s,a) - b(s)) \right ] 
\end{aligned}
\end{align}

As in previous work, an entropy-regularized policy gradient simply add the gradient of the entropy of the current policy with a tunable parameter $\alpha$ in order to encourage exploration. The entropy term is: 
%\begin{align}
%\begin{aligned}
\begin{equation}
\begin{split}
&\mathbb{E}_{s} \left [ \alpha \nabla_\theta H(\pi_Q(\cdot | s)) \right ] \\
=& \mathbb{E}_{s} \left [ \alpha \nabla_\theta \sum_a - \pi_Q(a | s) \log \pi_Q(a | s) \right ] \\
=& \mathbb{E}_{s} \left[ \alpha \sum_a - \nabla_\theta \pi_Q(a | s) \log \pi_Q(a | s) 
	\right. 
    \\
 	& \left. \qquad
    - \pi_Q(a | s) \nabla_\theta \log \pi_Q(a | s) \right] \\
=& \mathbb{E}_{s} \left [ \alpha \sum_a - \nabla_\theta \pi_Q(a | s) \log \pi_Q(a | s) 
	\right.
    \\
    & \left. \qquad
    -  \pi_Q(a | s) \frac{1}{\pi_Q(a | s)} \nabla_\theta \pi_Q(a | s) \right ] \\
=& \mathbb{E}_{s} \left [ \alpha \sum_a - \nabla_\theta \pi_Q(a | s) \log \pi_Q(a | s) \right ] \\
=& \mathbb{E}_{s} \left [ \alpha \sum_a - \pi_Q(a | s)  \nabla_\theta \log \pi_Q(a|s) \log \pi_Q(a | s) \right ] \\
\end{split}
\end{equation}

%\end{aligned}
%\end{align}

putting the two terms together and using the energy-based policy formulation (Eq.~\eqref{eq:14}) :

\begin{align*}
\begin{aligned}
&\mathbb{E}_{s,a} \left[ \nabla_\theta \log \pi_Q(a|s) (\hat{Q}(s,a) - b(s)) + \alpha \nabla_\theta H(\pi_Q(\cdot | s)) \right]  \\
=& \mathbb{E}_{s} \left [ \sum_a \pi_Q(s, a) \nabla_\theta \log \pi_Q(s, a) 
	\right.
    \\
    & \left. \qquad
	(\hat{Q}(s,a) - b(s) - (Q(s,a)-V_Q(s))) \right ] 
\end{aligned}
\end{align*}
If we let the baseline $b(s)$ be $V_Q(s)$, we get the update:
\begin{align}
\begin{aligned}
& \mathbb{E}_{s} \left [ \sum_a \pi_Q(s, a) \nabla_\theta \log \pi_Q(s, a) (\hat{Q}(s,a) - Q(s,a)) \right ] \\
= & \frac{1}{\alpha} \mathbb{E}_{s, a} \left [ (\nabla_\theta Q(s, a) - \nabla_\theta V_Q(s)) (\hat{Q}(s,a) - Q(s,a)) \right ] \\
\end{aligned}
\end{align}
where $\hat{Q}(s, a)$ could be obtained through bootstrapping by $R(s, a) + \gamma V_Q(s')$. In practice $V_Q(s')$ is computed from a target network. For the critic, the update is: 

\begin{align}
\begin{aligned}
&\mathbb{E}_s \left [ \nabla_\theta \frac{1}{2} (V_Q(s) - \hat{V}(s))^2  \right ] \\
=& \mathbb{E}_s \left [ \nabla_\theta V_Q(s) (V_Q(s) - \hat{V}(s)) \right ]
\end{aligned}
\end{align}
where $\hat{V}(s)$ could be similarly obtained by bootstrapping: $\hat{V}(s) = \mathbb{E}_a \left [ R(s, a) +\gamma V_Q(s') \right ] +  \alpha H(\pi_Q(\cdot|s))$.

\subsection{Effects of Imperfect Demonstrations}

\begin{figure*}[t]
    \centering
    \begin{subfigure}[t]{0.5\textwidth}
        \centering
        \includegraphics[width=\textwidth]{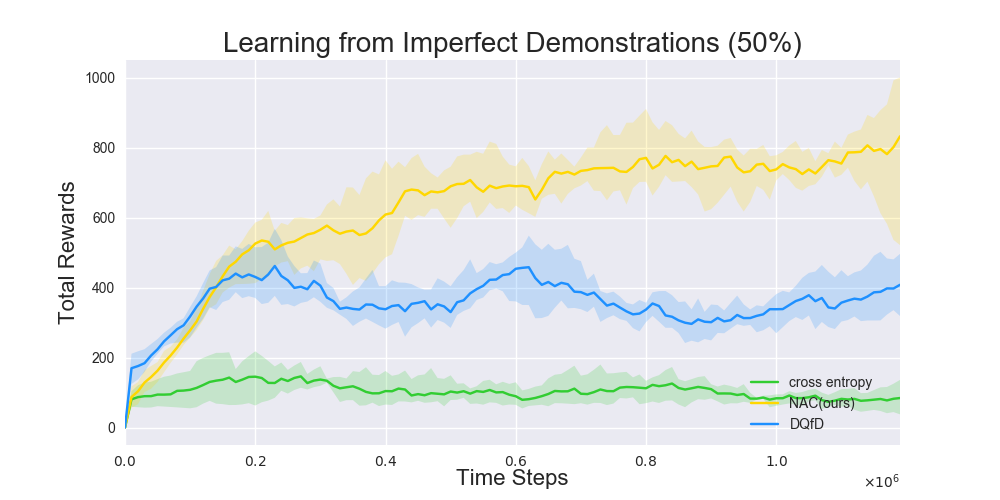}
    \end{subfigure}%
    ~
    \begin{subfigure}[t]{0.5\textwidth}
        \centering
        \includegraphics[width=\textwidth]{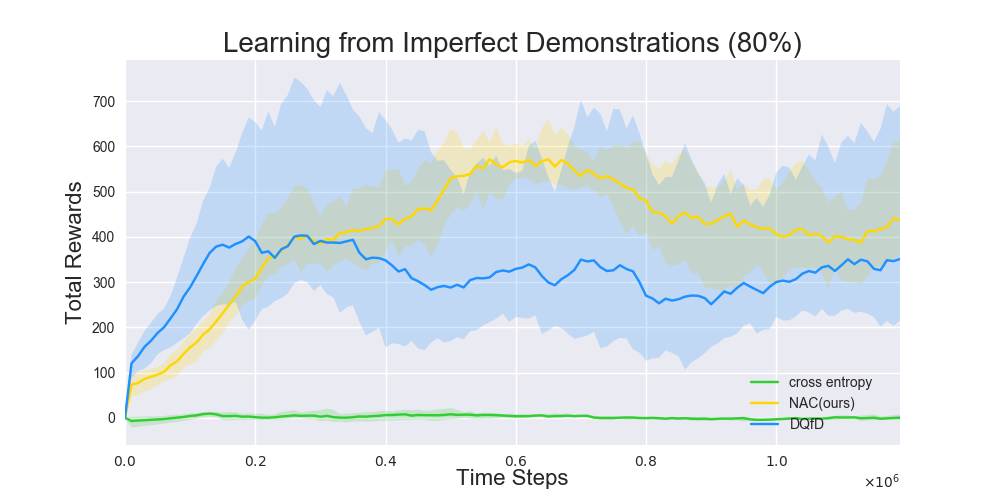}
    \end{subfigure}%  
    \caption{More results when introducing imperfect demonstrations. Left figure shows when there are 50\% imperfect actions and the right one shows the case for 80\%. \yang{Our NAC method is highly robust to noisy demonstrations. }}
    \label{fig:appendix_imperfect}
\end{figure*}

See Figure \ref{fig:appendix_imperfect} for more results for imperfect demonstrations when the amount of noise varies.

\subsection{Effects of Demonstration Amount}
\begin{figure*}[t]
    \centering
    \begin{subfigure}[t]{0.5\textwidth}
        \centering
        \includegraphics[width=\textwidth]{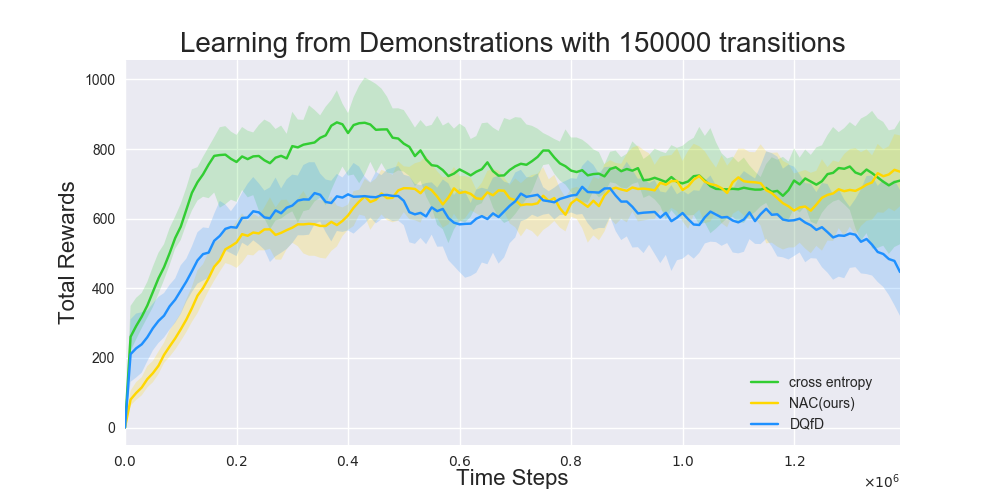}
    \end{subfigure}%
    ~
    \begin{subfigure}[t]{0.5\textwidth}
        \centering
        \includegraphics[width=\textwidth]{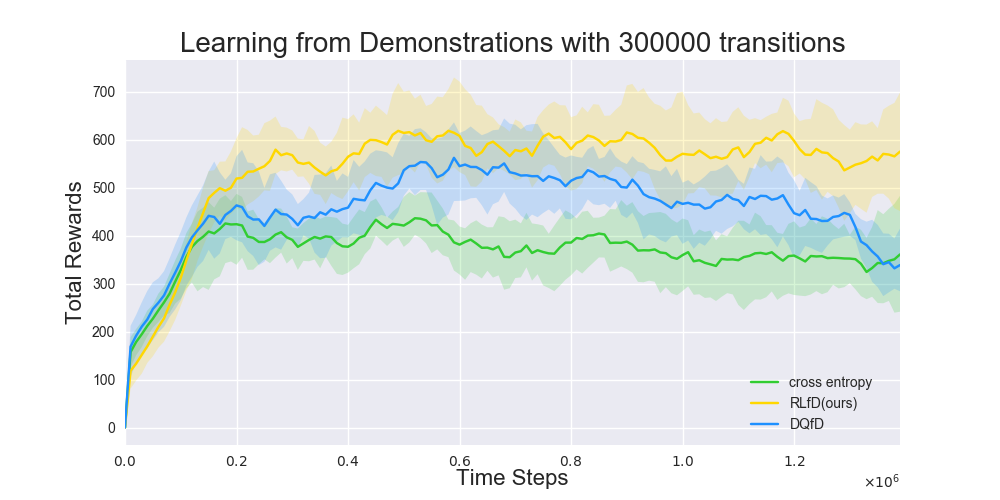}%{figures/lfsd300000}
      
    \end{subfigure}%  
    \caption{More results when varying the amount of demonstrations. The left and right figures show when there are 150k and 300k transitions respectively. \yang{Our NAC method achieves superior performance with a large amount of demonstrations and is comparable to supervise methods with smaller amount of demonstrations.} }
    \label{fig:appendix_amount}
\end{figure*}

See Figure \ref{fig:appendix_amount} for more results on the effect of the demonstration amount. 

\subsection{Experiment Details}

\textbf{Network Architecture:} We use the same architecture as in \cite{mnih2015human} to parametrize $Q(s, a)$. With this Q parametrization, we also output $\pi_Q(a|s)$ and $V_Q(s)$ based on Eq.~\eqref{eq:14} and Eq.~\eqref{eq:13}.

\textbf{Hyper-parameters:} We use a replay buffer with a capacity of 1 million steps and update the target network every 10K steps. Initially, the learning rate is linearly annealed from 1e-4 to 5e-5 for the first $1/10$ of the training process and then it is kept as a constant (5e-5). Gradients are clipped at 10 to reduce training variance. The reward discount factor $\gamma$ is set to 0.99. We concatenate the 4 most recent frames as the input to the neural network. For the methods with an entropy regularizer, we set $\alpha$ to 0.1, following \cite{schulman2017equivalence}. We truncate the importance sampling weighting factor $\beta =  \min \left \{ \frac{\pi_Q(a|s)}{\mu(a|s)}, c \right \}$ at 10, i.e., $c=10$.
\end{document}